# From Two-Stream to One-Stream: Efficient RGB-T Tracking via Mutual Prompt Learning and Knowledge Distillation

Luo Yang
The Aerospace Information Research Institute
Chinese Academy of Sciences
Beijing China
luoyang211@mails.ucas.ac.cn

Guo Xiqing[*]
The Aerospace Information Research Institute
Chinese Academy of Sciences
Beijing China
guoxq100036@aircas.ac.cn

Li Hao
The Aerospace Information Research Institute
Chinese Academy of Sciences
Beijing China
lihao231@mails.ucas.ac.cn

## ABSTRACT

Due to the complementary nature of visible light and thermal infrared modalities, object tracking based on the fusion of visible light images and thermal infrared images (referred to as RGB-T tracking) has received increasing attention from researchers in recent years. How to achieve more comprehensive fusion of information from the two modalities at a lower cost has been an issue that researchers have been exploring. Inspired by visual prompt learning, we design a novel two-stream RGB-T tracking architecture based on cross-modal mutual prompt learning, and use this model as a teacher to guide a one-stream student model for rapid learning through hierarchical distillation techniques. Extensive experiments have shown that, compared to similar RGB-T trackers, our designed teacher model achieved the highest precision rate, while the student model, with comparable precision rate to the teacher model, realized an inference speed more than three times faster than the teacher model.(Codes will be available if accepted.)

## CCS CONCEPTS

•Computing methodologies~Artificial intelligence~Computer vision~Computer vision problems~Tracking

## 1 INTRODUCTION

With the gradual maturation and prevalence of thermal infrared imaging devices, object tracking based on the fusion of visible light and thermal images (RGB-T tracking) is receiving increasing attention from researchers. Thermal infrared imaging, due to its insensitivity to lighting changes and ability to penetrate through fog and haze to a certain extent, can serve as a good complementary modality to visible light imaging. Conversely, the rich color and texture information present in visible light imaging is lacking in thermal infrared imaging. Therefore, fusing information from the two modalities for tracking can expand the application scenarios of algorithms and improve robustness. Early RGB-T tracking models were mostly anchor-based[1-12], with fixed aspect ratios for the generated bounding boxes, making them difficult to adapt to challenges such as severe target deformation. Consequently, in recent years, they have been gradually replaced by Transformer-based anchor-free solutions[13-18].

*Corresponding authors

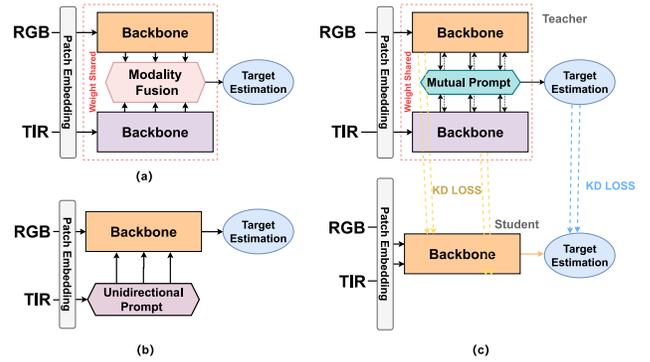

**Figure 1 Architectures of different RGB-T tracking models. (a) Two-stream structure. (b) One-stream structure(ViPT [41]). (c) Our proposed structure.**

Despite the generally high performance of current RGB-T models, the following challenges persist: i) Due to the high annotation cost of RGB-T tracking data (requiring special equipment and alignment of visible light and thermal infrared images), the existing related datasets[19-23] all suffer from insufficient volume. ii) Compared to single object tracking (SOT) tasks using only RGB images, RGB-T tracking models have a higher computational burden due to the addition of infrared images, necessitating a trade-off between precision rate and inference speed. iii) As the tracking environment changes, the dominant modality with an advantage may dynamically shift across different frames. How to adaptively identify the dominant modality during the tracking process, suppress noise from low-quality modalities, and fully leverage the complementary nature of the two modalities has remained a challenge.

Considering the issue of insufficient multi-modal data, mainstream RGB-T tracking methods typically initialize their backbones with pre-trained parameters from the SOT task. Additionally, some methods[40] have attempted to supplement the training data with infrared images synthesized from RGB images. However, since the synthesized data struggles to simulate the real-world scenario of modality dominance shifts, this approach has limited effectiveness in improving model performance.

In terms of model architecture, most current RGB-T tracking methods adopt a Siamese two-stream architecture[1, 10, 25, 26, 28, 40]. In this structure, two backbones separately extract features from the two modalities, which is very intuitive(As shown in Figure 1(a)). The advantage lies in reduced training difficulty and the ease of extracting features from different branches for fusion. However, the biggest issue with this structure is that the images from the two modalities need to go through separate forward processes, and an additional modality fusion module is required, greatly slowing down the inference speed. Early RGB-T tracking methods based on the two-stream architecture primarily employed complex attention mechanisms for modality fusion[10, 25, 40]. With the rise of visual prompt learning techniques, some methods[1, 28] have attempted to design lightweight prompt modules to achieve modality fusion and reduce the computational burden of the model. However, the impact on improving model efficiency is extremely limited. In their work ViPT[41], Zhu et al. adopted a single backbone for feature extraction and superimposed the auxiliary modality (infrared) information onto the input space of the dominant modality (RGB) through prompt learning, resulting in a significant efficiency boost(As shown in Figure 1(b)). However, this approach overlooked the dynamic switching of the dominant modality in multi-modal tracking scenarios and failed to effectively utilize the complementary information between modalities.

Based on the above analysis, we aim to design a simple and efficient one-stream RGB-T tracking architecture that completes feature extraction and deep fusion of the two modality images within a single Transformer backbone. In fact, in the field of single object tracking, previous works[3, 5, 33] have fully demonstrated that by simultaneously embedding the target template and search region and feeding it into a Transformer encoder, under proper constraints and with a large amount of labeled data for training, the encoder can learn the matching relationship between the template region and the search region. Quite naturally, in the RGB-T tracking domain, we wonder whether it is possible to simultaneously embed the visible light and thermal infrared modality images into a sequence and directly input it into a single Transformer backbone, leveraging the multi-head self-attention mechanism in the encoder layers to model the complementary relationship between the two modalities, thereby achieving modality fusion? Obviously, this approach has a major issue: since both modalities are input simultaneously, the token length doubles, and this increase in data dimension also means that during training, the Transformer model requires more data to effectively learn how to process these longer inputs, while the available RGB-T tracking data for training is insufficient. This leads to the model's inability to fully learn the complementary relationship between the two modalities and a risk of overfitting.

To address this problem, our solution is to first train an excellent two-stream RGB-T tracking teacher model, and then rapidly transfer the complementary modality information learned by the teacher to the one-stream student model through knowledge distillation(As shown in Figure 1(c)), thereby overcoming the issue of insufficient data and achieving a dual improvement in the student model's precision rate and real-time performance.

Specifically, our teacher model is based on the OSTrack[33], where we extend the backbone to a Siamese architecture (with shared weights). After separately embedding the images of the two modalities, we design a novel multi-modal mutual prompter. These prompters are inserted between every encoder layer of the two backbones, without distinguishing the dominant modality, to enable the network to adaptively determine the advantageous modality. Unlike the prompters in ViPT[41] that only compute weights within tokens, our designed prompters introduce attention operations on both the inter-token and intra-token (spatial) dimensions, learning the weights of each token from the two modalities as well as the weights of different spatial positions within each token. Furthermore, to enable multi-level information exchange and integration within the backbone, the input to each prompter includes not only the output of the previous encoder layer from the same modality branch, but also the output of the previous encoder layer from the other modality branch and the output of the prompter from the previous layer of the same modality branch. These three inputs are integrated through element-wise addition to obtain the prompter's output. The prompter's output is then superimposed on the output of the previous encoder layer, serving as the input of the next encoder layer. Finally, the features from the outputs of the two backbones are concatenated and dimensionality-reduced through a linear layer, before being fed into the localization head to regress the bounding box.

As for the student model, for the sake of simplicity, we directly adopted the same one-stream architecture as OSTrack[33], with the only difference being that we concatenated the tokens of the two modality images and input them together into the backbone. During the training process, the teacher model enables rapid knowledge transfer to the student through a hierarchical distillation strategy. Specifically, we apply constraints at two levels: the intermediate features of the backbone and the localization response maps, guiding the student model to mimic the feature distributions and response distributions of the teacher.

Evaluations and comparisons on public RGB-T datasets demonstrate that the student model obtained through our designed knowledge distillation process can achieve SOTA performance in terms of both precision rate and real-time capability, outperforming the vast majority of similar algorithms.

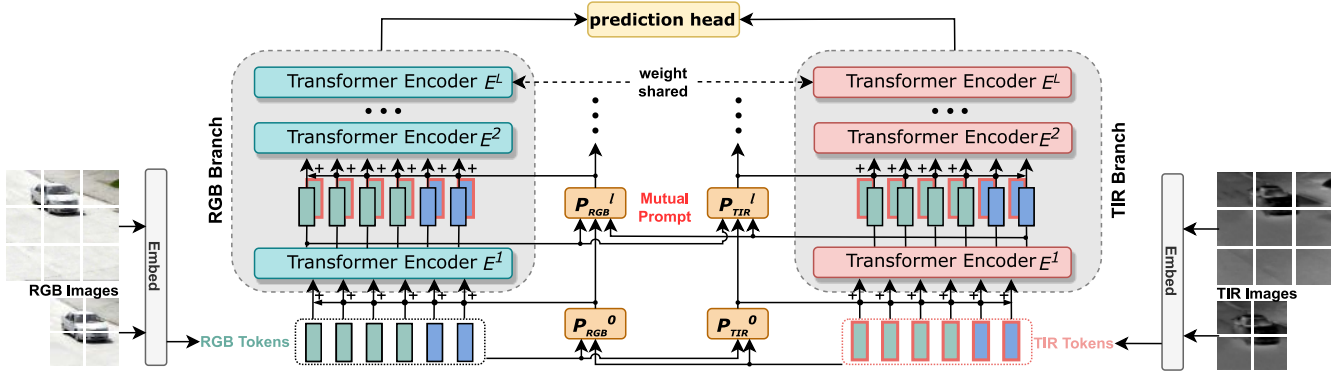

Figure 2 The overall architecture of our proposed teacher model, with the visible branch on the left and the thermal branch on the right. "+" stands for element-wise addition.

Meanwhile, it significantly outperforms the result of direct supervised training using RGB-T data in terms of precision rate.
Our contributions can be concluded as following aspects:
- Designed an RGB-T tracking architecture based on mutual prompt learning. By establishing bidirectional modal information interaction channels, it enables the complementary fusion of different modality images during the feature extraction stage, thereby realizing the model's adaptive judgment of the dominant modality.
- A more effective prompter called multi-modal mutual prompter (MMMP) was designed. It adaptively generates cross-modal weights through two attention mechanisms while taking into account historical information, thereby achieving high-quality information fusion.
- We designed a hierarchical knowledge distillation strategy from a two-stream structure to a one-stream structure. Through feature constraints and response constraints, it guides the student model to rapidly learn the complementary characteristics between the two modalities, thereby overcoming the training difficulties caused by the increase in token length. Significantly improved the inference speed.

## 2 RELATED WORK

### 2.1 Visual Prompt Learning

Prompt words, as a form of additional content added to text, have been widely applied in the NLP field to allow pre-trained models to better adapt to specific downstream tasks[6]. Recently, researchers have gradually begun exploring the introduction of prompt learning methods in the field of computer vision. VPT[11] was the earliest work to explore the feasibility of prompt learning in the vision domain, where freezing the backbone parameters and introducing a small number of learnable parameters into the input space achieved downstream performance comparable to full fine-tuning. ProTrack[32] was the first to introduce prompt learning into the multi-modal tracking field. Building upon this, ViPT[41] introduced a concise learnable prompter module, but failed to consider that the dominant modality may dynamically switch during tracking, resulting in an inability to fully leverage the complementary nature of the two modalities. BAT[1] inserted prompter modules into the encoder layers of a two-stream Transformer backbone, effectively adapting to the problem of drifting dominant modalities. However, similar to ViPT[41], its prompter only computed weights within the feature dimension of tokens, without considering the weights between tokens.

### 2.2 Knowledge Distillation Methods

Knowledge distillation was originally proposed by Hinton et al. [8] with the goal of transferring the complex knowledge structures within large models to smaller models through specific supervised training methods. In addition to detection[4] and segmentation[19] tasks, KD can also be applied to multi-modal tracking tasks. Zhang et al.[39] introduced knowledge distillation into the RGB-T tracking field, but their work focused on distilling independent two-stream networks into two-stream networks with shared parameters, resulting in limited improvement in inference speed. Wang et al.[30] introduced knowledge distillation to the RGB-E tracking field, transferring knowledge from the teacher model to a student model that only takes event stream data as input. Since the student model lost the visible light modality input, the complementary information between modalities was also lost, causing a more significant performance degradation in the student model.

## 3 METHOD

This chapter will provide a detailed introduction to our proposed RGB-T tracking method based on mutual prompt learning, and introduce how to transfer the knowledge from the two-stream teacher model to the one-stream student model through hierarchical distillation.

### 3.1 Baseline Tracking Process

The basic form of single-object tracking (SOT) involves taking the target region from the initial frame as the template, denoted as $Z_{RGB} \in \mathbb{R}^{H_z \times W_z \times 3}$, and searching for the template in the subsequent frame $X_{RGB} \in \mathbb{R}^{H_x \times W_x \times 3}$ to locate the target by bounding it.

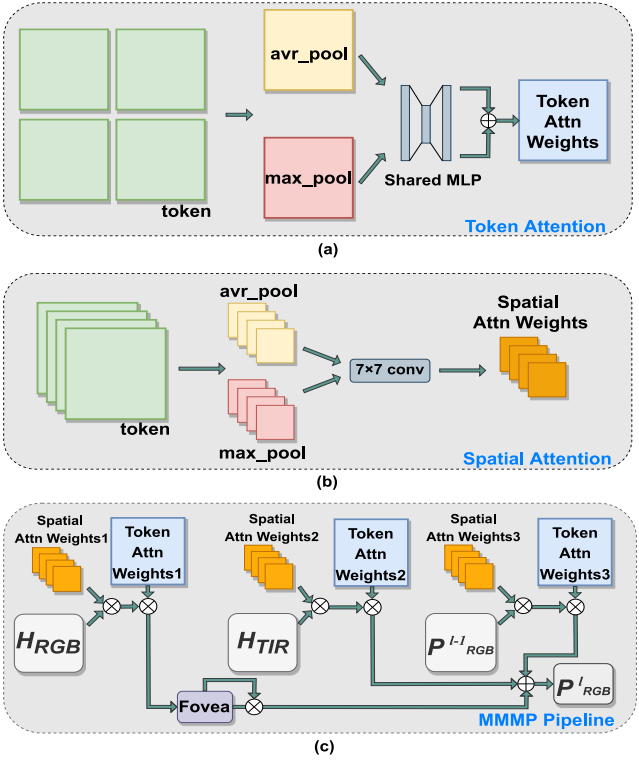

Figure 3. The overall architecture of Multi-Modal Mutual Prompter (MMMP), (a) is the process of generating token attention weights, (b) is the process of generating spatial attention weights, (c) is the overall pipeline.

For Transformer-based SOT models like the baseline model [33], the first step is to convert $Z_{RGB}, X_{RGB}$ into patches of size $P \times P$ through embedding (In ViT-B, $P = 16$):

$$\{Z_{RGB}, X_{RGB}\} \to \{Z_{RGB}^P \in \mathbb{R}^{N_z \times D}, X_{RGB}^P \in \mathbb{R}^{N_X \times D}\} \quad (1)$$

Where $N_z = \frac{H_z W_z}{P^2}$, $N_X = \frac{H_x W_x}{P^2}$, $D = P^2 \times C$, ($C$ is the number of image channels, here is 3). Next, $Z_{RGB}^P$, $X_{RGB}^P$ are concatenated and fed into the backbone to learn features and facilitate interaction between the template and search regions:

$$H_{RGB} = Concat(Z_{RGB}^P, X_{RGB}^P) \quad (2)$$

$$B_{RGB} = h(f(H_{RGB})) \quad (3)$$

Where $H_{RGB} \in \mathbb{R}^{N \times D}, N = N_X + N_Z$, $f$ represent the ViT backbone, $h$ represent the localization head, $B_{RGB}$ represent the final output bounding box.

### 3.2 Teacher Model Design

*3.2.1 Overview.* As shown in Figure 2, in teacher model, in addition to the visible modality, we also introduce the thermal modality. Therefore, we have:

$$\{Z_{TIR}, X_{TIR}\} \to \{Z_{TIR}^P \in \mathbb{R}^{N_z \times D}, X_{TIR}^P \in \mathbb{R}^{N_X \times D}\} \quad (4)$$

$$H_{TIR} = Concat(Z_{TIR}, X_{TIR}) \quad (5)$$

Moreover, we extend the backbone into a two-stream architecture, where $f_{RGB}$ represents the backbone for extracting RGB features, and $f_{TIR}$ represents the backbone for extracting thermal features. Let $E_{RGB}^l$ denote the encoder in the l-th layer of $f_{RGB}$, and $E_{TIR}^l$ denote the encoder in the l-th layer of $f_{TIR}$ (with a total of L layers). $H_{RGB}^l$ represents the output of $E_{RGB}^l$, and $H_{TIR}^l$ represents the output of $E_{TIR}^l$, then we have:

$$\begin{aligned} H_{RGB}^l &= E_{RGB}^l(H_{RGB}^{l-1} + P_{RGB}^l) \\ P_{RGB}^l &= \mathbf{M}(H_{RGB}^{l-1}, P_{RGB}^{l-1}, H_{TIR}^{l-1}) \\ H_{TIR}^l &= E_{TIR}^l(H_{TIR}^{l-1} + P_{TIR}^l) \\ P_{TIR}^l &= \mathbf{M}(H_{TIR}^{l-1}, P_{TIR}^{l-1}, H_{RGB}^{l-1}) \end{aligned} \quad (6)$$

Where $l = 1,2,...,L$, $\mathbf{M}(*)$ represents the Multi-Modal Mutual Prompter(MMMP), and $P_{RGB}^l$ and $P_{TIR}^l$ represent the output of the MMMP in the l-th layer. Specifically, when l=0, we have $P_{RGB}^0 = \mathbf{M}(H_{RGB}^0, H_{TIR}^0)$ and $P_{TIR}^0 = \mathbf{M}(H_{TIR}^0, H_{RGB}^0)$.

Lastly, the features from the two branches, after mutual prompt learning, are concatenated. Then, they are passed through a linear layer to reduce the channel dimensionality before being fed into the localization head for classification prediction and target box regression:

$$B = h\left(DR\left(Concat(H_{RGB}^L, H_{RGB}^L)\right)\right) \quad (7)$$

Where $DR$ represents the linear layer for dimensionality reduction. The details of the localization head $h$ can be referred to OSTrack[33].

*3.2.2 Multi-Modal Mutual Prompter.* We obtained insights on prompter design from VPT[11] and made improvements based on that. As shown in Equation 6, the MMMP module has three input branches: the output of the encoder of the current modality from the previous layer, the output of the encoder of the other modality from the previous layer, and the output of the previous MMMP. As shown in Figure 3, The workflow of the MMMP module consists of three steps:

(1) Spatial attention operations are performed on each of the three input branches. Specifically, for a particular branch, let's assume the input tokens are from RGB branch, denoted as $H_{RGB} \in \mathbb{R}^{N \times D}$, average pooling and max pooling are first applied:

$$W_{avr}^s = mean(H_{RGB}, dim \sim D) \quad (8)$$

$$W_{max}^s = max(H_{RGB}, dim \sim D) \quad (9)$$

Next, $W_{avr}^s$ and $W_{max}^s$ undergo channel dimension reduction in the N dimension using a 1×1 convolutional layer $g_{s1}$. This projects the features into a lower-dimensional latent embedding. The resulting features are then passed through a ReLU layer for non-linear enhancement before being projected back to the original dimension using another 1×1 convolutional layer $g_{s2}$. This generates a weight map with the same size as the token dimension $D$:

$$W_{spatial} = g_{s2}\left(relu(g_{s1}(W_{avr}^s))\right) + g_{s2}\left(relu(g_{s1}(W_{max}^s))\right) \quad (10)$$

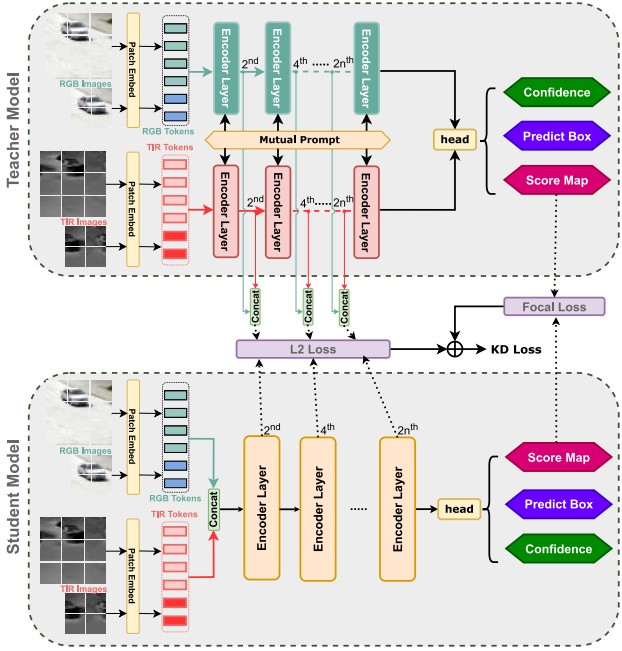

**Figure 4. Architecture diagram of hierarchical distillation. To simplify the presentation, the Multi-Modal Mutual Prompter in the teacher model is omitted.**

Finally, $W_{spatial}$ is multiplied element-wise with the original tokens to obtain the tokens with redistributed weights in the spatial dimension:

$$H_{RGB}^{s} = H_{RGB} * W_{spatial} \quad (11)$$

(2) Token attention operations are performed on each of the three input branches. Specifically, for a particular branch, the average and maximum values are computed along the $N$ dimension of the tokens:

$$W_{avr}^{t} = mean(H_{RGB}^{s}, dim\sim N) \quad (12)$$

$$W_{max}^{t} = \max(H_{RGB}^{s}, dim\sim N) \quad (13)$$

Then, $W_{avr}^{t} \in \mathbb{R}^{1 \times D}$ and $W_{max}^{t} \in \mathbb{R}^{1 \times D}$ are concatenated along the first dimension, and a 7×7 convolutional layer $g_t$ with padding is used to reduce the first dimension back to 1, resulting in a $D$-dimensional weight map $W_{token}$:

$$W_{token} = g_t \cdot Concat(W_{avr}^{t}, W_{max}^{t}) \quad (14)$$

Then, the weight map is multiplied element-wise with the input tokens to obtain the token sequence with reweighted values:

$$H_{RGB}^{t} = H_{RGB}^{s} * W_{token} \quad (15)$$

(3) Execute the spatial fovea operation on the token of current modality, which first applies a λ-smoothed spatial softmax across all the spatial dimensions, and produces the enhanced embeddings by applying the channel-wise spatial attention-like mask $W_{fovea}$ over $H_{RGB}^{t}$. Then, the tokens from the other two branches are added to obtain the output of MMMP：

$$P_{RGB}^{l} = W_{fovea} \cdot H_{RGB}^{t} + H_{TIR}^{t} + P_{RGB}^{l-1\ t} \quad (16)$$

Where $H_{TIR}^{t}$ represent the enhanced tokens from TIR branch after the two attention operations, and $P_{RGB}^{l-1\ t}$ represents the output of the previous MMMP after the two attention operations. The calculation of the mutual prompt of the TIR branch is the same, only need to exchange the modality labels.

### 3.3 Hierarchical Distillation Process

This section will detail the hierarchical distillation process from two aspects: feature-based distillation design and response-based distillation design(As shown in Figure 4). Additionally, for the structure of the one-stream student model, please refer to OS-Track[33].

*3.3.1 Feature based Distillation.* Let the output features of the l-th encoder layer of the teacher model be $H_t^l$ (where the features of the visible light branch are denoted as $H_{t_{RGB}}^l$, and the features of the thermal infrared branch are denoted as $H_{t_{TIR}}^l$). The output features of the l-th encoder layer of the student model are $H_s^l$. Since the intermediate layer features of the student model are exactly twice the token dimension of the intermediate layer features of the teacher model, when designing the distillation loss function, we can directly concatenate the features of the two branches from the teacher model to guide the training of the student model, without the need for additional linear transformation operations. Let the concatenated features of the teacher model be $H_{t_{Fus}}^l$, then we have:

$$H_{t_{Fus}}^l = Concat(H_{t_{RGB}}^l,\ H_{t_{TIR}}^l,\ dim\sim N) \quad (17)$$

To enable the student model to rapidly learn the parameter distributions within the teacher model under the guidance of the teacher's intermediate layer features, we design the intermediate feature-based distillation loss $\mathcal{L}_{MF}$ as follows:

$$\mathcal{L}_{MF} = \sum_{i=1}^{n}\left(2i * \mathcal{L}_2(H_{t_{Fus}}^i,\ H_s^i)\right) \quad (18)$$

Where n is half of the number of encoder layers, which is 6 in ViT-B, and $\mathcal{L}_2(*)$ represents the L2 loss function. The reason for extracting intermediate layer features at intervals is to consider that excessive constraints may have a negative impact on the generalization performance. The effectiveness of this design will be validated in the subsequent experimental section.

*3.3.2 Response based Distillation.* The response map is the probability map generated in the target localization head, used for bounding box regression. Naturally, we would think that if the response map generated by the student model can maximally mimic that of the teacher model, then the student model's tracking performance can approach the teacher model to the greatest extent. Here, let the response map generated by the teacher model's localization head be $R_t$, and the response map generated by the student model's localization head be $R_s$. First, we normalize $R_t$ and $R_s$ by dividing them by a temperature coefficient τ (empirically set to 2), and then use the focal loss[13] to calculate the similarity between the response maps, as expressed by the following formula:

$$\mathcal{L}_{RM} = \mathcal{L}_{focal}(R_t/\tau,\ R_s/\tau) \quad (19)$$

| Method | Backbone | Pub | GTOT | | RGBT234 | | LasHeR | | VTUAV-ST | | VTUAV-LT | | FPS |
|---|---|---|---|---|---|---|---|---|---|---|---|---|---|
| | | | PR | SR | PR | SR | PR | SR | PR | SR | PR | SR | |
| FSRPN[12] | ResNet-50 | ICCVW'19 | 89.0 | 69.5 | 71.9 | 52.5 | - | - | 65.3 | 54.4 | 36.6 | 31.4 | 36.8 |
| mfDiMP[35] | ResNet-50 | ICCVW'19 | 83.6 | 69.7 | 84.6 | 59.1 | 44.8 | 34.3 | 67.3 | 55.4 | 31.5 | 27.2 | 34.6 |
| DAFNet[7] | VGG-M | ICCVW'19 | 89.1 | 71.6 | 79.6 | 54.4 | 44.8 | 31.1 | 62.0 | 45.8 | 25.3 | 18.8 | 20.5 |
| DAPNet[42] | VGG-M | ACM MM'19 | 88.2 | 70.7 | 76.6 | 53.7 | 43.1 | 31.4 | - | - | - | - | - |
| MANet[20] | VGG-M | TIP'20 | 89.4 | 72.4 | 77.7 | 53.9 | 45.5 | 32.6 | - | - | - | - | 2.1 |
| CAT[16] | VGG-M | ECCV'20 | 88.9 | 71.7 | 80.4 | 56.1 | 45.0 | 31.4 | - | - | - | - | - |
| CMPP[27] | VGG-M | CVPR'20 | **92.6** | 73.8 | 82.3 | 57.5 | - | - | - | - | - | - | - |
| JMMAC[37] | VGG-M | TIP'21 | 90.2 | 73.2 | 79.0 | 57.3 | - | - | - | - | - | - | - |
| MANet++[22] | VGG-M | TIP'21 | 88.2 | 70.7 | 79.5 | 55.9 | 46.7 | 31.4 | - | - | - | - | 25.4 |
| ADRNet[36] | VGG-M | IJCV'21 | 90.4 | 73.9 | 80.7 | 57.1 | - | - | 62.2 | 46.6 | 23.5 | 17.5 | 25.0 |
| SiamCDA[40] | ResNet-50 | TCSVT'21 | 87.7 | 73.2 | 79.5 | 54.2 | - | - | - | - | - | - | 24.0 |
| TFNet[44] | VGG-M | TCSVT'22 | 88.6 | 72.9 | 80.6 | 56.0 | - | - | - | - | - | - | - |
| DMCNet[23] | VGG-M | TNNLS'22 | - | | 83.9 | 59.3 | 49.0 | 35.5 | - | - | - | - | - |
| MFGNet[29] | VGG-M | TMM'22 | 88.9 | 70.7 | 78.3 | 53.5 | - | - | - | - | - | - | 3.0 |
| APFNet[31] | VGG-M | AAAI'22 | 90.5 | 73.7 | 82.7 | 57.9 | 50.0 | 36.2 | - | - | - | - | 1.9 |
| HMFT[38] | ResNet-50 | CVPR'22 | 91.2 | 74.9 | 78.8 | 56.8 | - | - | 75.8 | 62.7 | 41.4 | 35.5 | 30.2 |
| HMFTLT[38] | ResNet-50 | CVPR'22 | - | - | - | - | - | - | - | - | 53.6 | 46.1 | 8.1 |
| MIRNet[9] | VGG-M | ICME'23 | 90.9 | 74.4 | 81.6 | 58.9 | | | - | - | - | - | 30.0 |
| ECMD[39] | ResNet-50 | CVPR'23 | 90.7 | 73.5 | 84.4 | 60.1 | | | | | | | 30.0 |
| ViPT[41] | ViT-B | CVPR'23 | - | - | 83.5 | 61.7 | 65.1 | 52.5 | - | - | - | - | 50.9* |
| TBSI[10] | ViT-B | CVPR'23 | - | - | 87.1 | 63.7 | 70.5 | 56.3 | - | - | - | - | 36.2* |
| MACFT[25] | ViT-B | Sensors'23 | 90.0 | 72.7 | 85.7 | 63.2 | 65.3 | 51.4 | 80.1 | 66.8 | 54.1 | 46.7 | 33.3* |
| TATrack[28] | ViT-B | Arxiv'24 | - | - | 87.2 | 64.4 | 70.2 | 56.1 | - | - | - | - | 26.1* |
| STMT[26] | ViT-B | Arxiv'24 | - | - | 86.5 | 63.8 | 67.4 | 53.7 | - | - | - | - | 39.1* |
| **Ours-Teacher** | ViT-B | - | **92.6** | **77.5** | **88.3** | **66.1** | 71.2 | 56.5 | **82.2** | **69.5** | **64.0** | **54.8** | 22.8* |
| **Ours-Student** | ViT-B | - | 92.4 | 77.3 | 87.3 | 65.1 | **71.4** | **56.7** | 80.2 | 67.4 | 63.6 | 54.0 | **81.8*** |

Table 1 Comparison with state-of-the-art RGBT tracking methods on multiple RGB-T datasets. "*" means that the inference speed of this method has been tested on the same GPU as the method proposed in this article.

## 4 EXPERIMENTS

### 4.1 Training and Inference Details

*4.1.1 Loss Function.* For the teacher model, its training adopts an end-to-end form. A total of three loss functions are used in the offline training stage, namely L1 loss, generalized GioU loss[2] and Focal Loss[13]. The overall loss function is expressed as follows (For specific implementation, please refer to OSTrack[33]):

$$\mathcal{L}_t = \mathcal{L}_{focal} + \lambda_{giou}\mathcal{L}_{giou} + \lambda_{L_1}\mathcal{L}_1 \quad (20)$$

For the student model, in addition to the above three types of loss functions, it also includes the two distillation losses mentioned in Section 3.3. The overall loss function is expressed as follows:

$$\mathcal{L}_s = \mathcal{L}_{focal} + \lambda_{giou}\mathcal{L}_{giou} + \\ \lambda_{L_1}\mathcal{L}_1 + \lambda_{RM}\mathcal{L}_{RM} + \lambda_{MF}\mathcal{L}_{MF} \quad (21)$$

*4.1.2 Parameter Details.* During training, 60,000 images were sampled in each epoch. The batchsize is set to 24 when training the teacher model and to 12 when distilling the student model. The teacher model was trained for a total of 15 epochs, and the student model was trained for 13 epochs. The learning rate for the backbone is set to 7.5e-5 while for other parts is set to 7.5e-4, and decays by 10 times after the 10th epoch. We use the AdmW[21] optimizer for iteration with a weight decay of 1e-4. The input image sizes for the network are as follows: the search frame is 256×256, and the template frame is 128×128. In terms of hyperparameters, $\lambda_{giou} = 2$, $\lambda_{L_1} = 5$, $\lambda_{RM} = 0.7$, $\lambda_{MF} = 0.035$. Additionally, for evaluation metrics, we use commonly used precision/success rate(PR/SR) metrics(by default, the best value of the two modalities is taken.) and set the center location error (CLE) threshold to the conventional value of 20 pixels. In particular, all the following experimental results were completed on a workstation with two RTX3090 GPUs, and all models were trained using the training set of the LasHeR dataset.

*4.1.3 Inference.* The tracking models designed in this article are all end-to-end inferences, with no additional online update operations.

### 4.2 Evaluate and Compare

*4.2.1 Public RGB-T Dataset Evaluation.* We evaluated our proposed method on four datasets: GTOT[14], RGBT234[15], LasHeR[17], and VTUAV[38]. VTUAV is divided into the short-term subset VTUAV-ST and the long-term subset VTUAV-LT. The results of the evaluation are shown in Table 1.

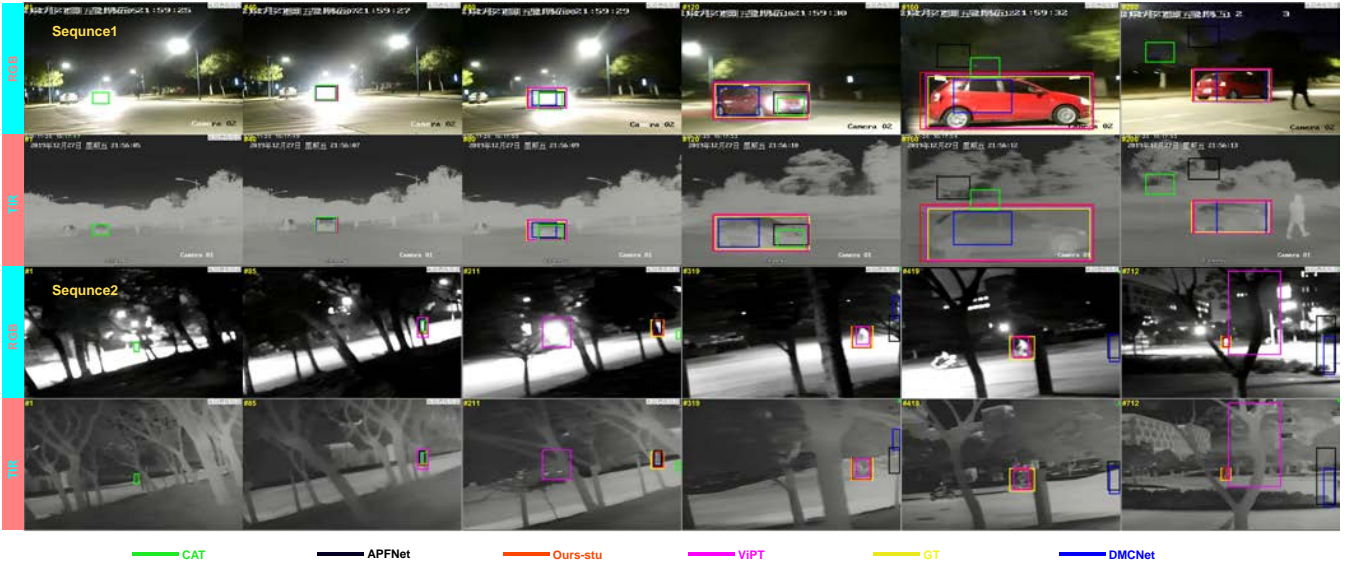

Figure 5. Qualitative comparison of our proposed method. From left to right, the time corresponds from first to last

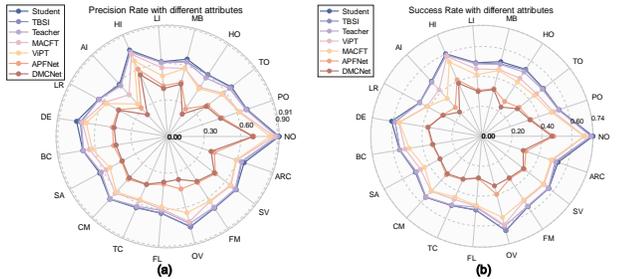

**Figure 6 Attribute-based evaluation on LasHeR dataset. (a) Precision Rate with different attributes. (b) Success Rate with different attributes.**

In the evaluation on the GTOT[14] dataset, our proposed teacher model achieved a precision rate of 92.6% and a success rate of 77.5%, outperforming most RGB-T trackers. The distilled student model, achieved a precision rate of 92.4% and a success rate of 77.3%, only 0.2% lower than the teacher model, and still outperformed most RGB-T trackers. In the evaluation on the RGBT234[15] dataset, our proposed teacher model achieved a precision rate of 88.3% and a success rate of 66.1%, also realizing SOTA performance. Compared to the teacher model, the student model only showed a 1% decrease in PR/SR metrics. On the more challenging LasHeR[17] dataset, our trained student model even slightly outperformed the teacher model, achieving a precision rate of 71.4% and a success rate of 56.7%, leading most RGB-T trackers. The reasons are analyzed as follows: **First**, since the student model is not significantly smaller than the teacher model in terms of parameters, the upper limit of the fitting ability will not be lower than the teacher model. **Secondly**, when the RGB-T data volume itself is not large, Under this circumstance, the complex fusion structure of the teacher model has a certain possibility of overfitting, so it is possible that the student model performs better than the teacher model.

It is also worth noting that although our model was only trained on the LasHeR dataset and was not specifically optimized for long-term tracking, both our teacher and student models significantly outperformed other algorithms on the long-term tracking subset of VTUAV[38], indicating that our designed tracker has good potential for long-term tracking to some extent.

*4.2.2 Attribute-Based Evaluation.* To further analyze the effectiveness of our proposed method, we evaluated it on 19 challenge attributes provided by the LasHeR dataset, and the results are shown in Figure 6. In terms of PR/SR, our student model performs very well under each challenge attribute, and surpasses the teacher model in quite a few attributes.

To analyze the performance of the student model in detail, we selected 6 representative challenging attributes for comparison. The first two are challenges unique to RGB-T tracking: thermal crossover and high illumination. The latter 4 are common challenges in single object tracking: fast motion, similar appearance, deformation, and total occlusion. We can observe that our student model performs excellently on these typical challenging attributes, outperforming the outstanding ViT-based trackers TBSI and ViPT, and even surpassing the teacher model. This further illustrates that our tracker can exploit the complementary properties between modalities very effectively.

| Attributes | Teacher | ViPT | TBSI | Student |
|---|---|---|---|---|
| TC | 50.4% | 46.0% | 50.1% | **50.6%** |
| HI | 59.2% | 54.2% | 58.2% | **60.2%** |
| FM | 56.0% | 51.5% | 55.7% | **56.2%** |
| SA | 49.6% | 46.5% | 50.2% | **50.5%** |
| DEF | 58.7% | 55.8% | 58.7% | **61.1%** |
| TO | 51.7% | 46.2% | 51.0% | **52.0%** |

**Table 2 Evaluation under six tough challenge attributes.**

*4.2.3 Qualitative Evaluation.* To provide a visual comparison, we selected two representative video sequences from the LasHeR dataset and visualized the tracking results of our proposed tracker

(student model) and several other similar RGB-T trackers(As shown in Figure 5). Sequence 1 includes challenges such as strong light interference and severe object deformation, while Sequence 2 includes challenges such as low light, low resolution, and occlusion.

Observing the results of Sequence 1, we can see that when the visible light modality target was not visible in the initial frame, our tracker effectively utilized the advantage of the infrared modality for tracking. Later on, it switched to the visible light modality as the dominant one, effectively addressing the challenge of severe object deformation.

In Sequence 2, we can observe that under dark conditions and with occlusion, our tracker was able to recover tracking after a brief adjustment, while most of the other trackers completely lost the target in the later stages. This reflects, to some extent, the strong robustness of our proposed method.

## 4.3 Ablation Studies

In this section, each component of our proposed method will be analyzed separately to validate their effectiveness.

*4.3.1 Teacher Model Component Analysis.* In this section, we conducted an ablation analysis on the components of the teacher model, setting up 5 control groups. As shown in Table 3, the first row represents the baseline model(finetuned in LasHeR dataset), the second row removes the spatial attention computation, the third row removes the token attention computation, the fourth row cancels the cross-layer connections between prompters, and the fifth row is the complete teacher model. The evaluation on the LasHeR dataset shows that the components we designed improved the performance of the tracker to varying degrees.

| Spatial Attn | Token Attn | History Information | PR | SR |
|---|---|---|---|---|
| | | | 61.0 | 48.3 |
| | √ | √ | 67.3 | 53.5 |
| √ | | √ | 67.4 | 53.6 |
| √ | √ | | 71.0 | 56.4 |
| √ | √ | √ | **71.2** | **56.5** |

**Table 3** Impact of different components on the performance of the teacher model (evaluate on the LasHeR dataset)

| Response KD | Feature KD | PR | SR |
|---|---|---|---|
| | | 75.6 | 55.7 |
| √ | | 82.9 | 62.0 |
| | √ | 86.9 | 64.7 |
| √ | √ | **87.3** | **65.1** |

**Table 4** Impact of different levels of distillation strategies on tracking performance (evaluate on the RGBT234 dataset)

*4.3.2 Distillation Method Analysis.* In this section, we experimentally tested the impact of each distillation strategy on the final tracking performance (evaluated on the RGBT234 dataset). As shown in Table 4, the first row represents not using any distillation strategy and only using the default loss function in the baseline model; the second row represents using only the response-based distillation strategy; the third row represents using only the feature-based distillation strategy; and the last row represents using both, which is the complete hierarchical distillation strategy. According to the evaluation results, we can conclude that different distillation strategies can effectively improve the RGB-T tracking performance, and the distillation strategy based on feature constraints provides a more significant improvement to the student model than the strategy based on response maps.

*4.3.3 Feature Combinations Analysis.* In this section, we verified the impact of extracting intermediate layer features in different combinations during the knowledge distillation process on the training results of the student model. As shown in Table 5, we set up four groups of experiments: (1) Extract the first six layers of features from the teacher model. (2) Extract the last six layers from the teacher model. (3) Extract all layer features from the teacher model. (4) Extract the even-numbered layer features from the teacher model.

The evaluation results on the LasHeR dataset show that using the even-numbered layer features of the teacher model to supervise the student model's learning can achieve better tracking performance. This may be because the student model needs to learn different levels of feature expressions, from low to high layers, from the teacher model. Learning only low-level or only high-level features is not sufficient. As for why using all layer features for constraint is not as good as ignoring some layers, it can be explained that the student model relies on the attention operation in the encoder layers for fusing different modalities, and there are still structural and logical differences from the teacher model. If the student model is required to adjust according to the feature structure of the teacher model, it may backfire.

| Combination method | PR | SR |
|---|---|---|
| First 6 layers | 70.6 | 56.1 |
| Last 6 layers | 70.8 | 56.2 |
| Full layers | 69.7 | 55.4 |
| Even-numbered layers | **71.4** | **56.7** |

**Table 5** Impact of different feature combinations in distillation on tracking performance (evaluate on the LasHeR dataset)

## 5 CONCLUSION

In this work, we proposed an RGB-T tracking architecture based on mutual prompt learning. By introducing information from the two modalities and considering hierarchical information through our designed prompters in an equivalent manner, the model can understands the complementary characteristics between modalities, effectively addressing the challenge of dominant modality drift during the tracking process. Furthermore, to simplify the tracking model and accelerate model inference, we proposed using knowledge distillation to compress the RGB-T tracking model from a two-stream structure to a one-stream, inherently learning the complementary characteristics between modalities within the Transformer encoder. This effectively overcomes the difficulty of training a single Transformer backbone due to the insufficient RGB-T data volume. Extensive experiments demonstrate that our proposed method achieves state-of-the-art performance, and we believe this method can be generalized to more other modalities.

# ACKNOWLEDGMENTS

# Supplementary Material

Luo Yang
The Aerospace Information Research Institute
Chinese Academy of Sciences
Beijing China
luoyang211@mails.ucas.ac.cn

Guo Xiqing[*]
The Aerospace Information Research Institute
Chinese Academy of Sciences
Beijing China
guoxq100036@aircas.ac.cn

Li Hao
The Aerospace Information Research Institute
Chinese Academy of Sciences
Beijing China
lihao231@mails.ucas.ac.cn

In the supplementary material, we will provide more experimental and evaluation details that are not shown in the main text, conduct additional ablation analysis, and provide more visualization results for reference to support our conclusions.

## A MORE ABLATION ANALYSIS

In this section, we analyze the impact of various components in the proposed teacher model on the inference speed. On this basis, we further analyze why there are significant differences in inference speed among similar one-stream tracking models.

### A.1 Inference Speed Analysis

In the encoder layers of the Transformer, the formula for computing the self-attention mechanism is as follows:

$$Attention\ (Q, K, V) = softmax\ (\frac{QK^T}{\sqrt{d_k}})V \quad (1)$$

Assuming the input token dimensions are (B, N, D), it is easy to derive that the computational complexity of the self-attention calculation (or the number of matrix multiplications) is:

$$\Omega\ (SA) = 4ND^2 + 2(N)^2 D \quad (2)$$

Assuming we only input the RGB modality, since the input search region size is 256×256×3, the template region size is 128×128×3, and the patch size is 16, we can obtain N=320, D=768. Similarly, when inputting two modalities simultaneously, the token length N would become 640. Substituting these two sets of data into equation 2, we can find that the computational complexity of inputting two modalities simultaneously is about 2.3 times that of inputting only one modality. This implies that in the case of adopting a two-stream structure where each branch processes only one modality, the model complexity should be close to the case of adopting a one-stream branch and processing two modalities in one branch. The inference speed data of the teacher model after removing all data fusion modules in Table 1 also proves this point.

Although theoretically, adopting a two-stream structure to process the two modalities separately and a one-stream structure to process the two modalities simultaneously are comparable in terms of computational complexity under the current input image size, in practice, our teacher model is significantly slower than the student model in terms of inference speed. Furthermore, the inference speed of the ViPT[15] model, which adopts a one-stream structure and only inputs a token length equivalent to a single modality, is also slower than our student model. Therefore, we can preliminarily conclude that the MLP/Adaptor structure inserted in the Transformer encoder layers has a significant impact on the model's inference speed.

Further analysis reveals that the inference speed of the model is also related to the specific implementation form of the MLP/Adaptor structure. In our teacher model and ViPT, the implementation of the prompt encoder adopts nn.Conv2D based on the cross-correlation operator to replace nn.Linear based on matrix multiplication. Due to the differences in underlying implementations and the different initialization mechanisms of Pytorch[5] for these two classes, researchers have found that using nn.Conv2D instead of nn.Linear generally achieves better fitting effects in actual experiments. However, the drawback is that its inference speed is slower than nn.Linear. The comparison between the two implementation methods is shown in Table 2.

Although the prompter is changed to be implemented based on nn.Linear, the inference speed of ViPT is still far lower than the theoretical value it should achieve. It is speculated that this might be due to the insertion of the prompter disrupting the parallel computing conditions originally present in the encoder layers.

Therefore, the inference speed advantage of our designed student model is more pronounced due to the elimination of all additional modal fusion structures. In fact, since there is still a considerable degree of redundant information between modalities, theoretically, in a one-stream architecture like the student model, we can further remove those redundant tokens to accelerate model inference, which we will leave for future work.

| Spatial Attn | Token Attn | History Information | FPS |
|---|---|---|---|
| √ | √ | √ | 22.8 |
| √ | √ |   | 33.4 |
| √ |   | √ | 33.0 |
|   | √ | √ | 41.9 |
|   |   |   | **78.8** |

**Table 1 Impact of different components on the inference speed of the teacher model.**

| One-Stream Tracker | FPS | PR |
|---|---|---|
| Student | **81.8** | **71.4** |
| Teacher-Conv2D | 22.8 | 71.2 |
| Teacher-Linear | 31.3 | 69.6 |
| ViPT-Conv2D[15] | 50.9 | 65.1 |
| ViPT-Linear[15] | 77.1 | 64.9 |

**Table 2 Impact of using nn.Conv2D and nn.Linear on inference speed and precision rate (evaluated on the LasHeR dataset).**

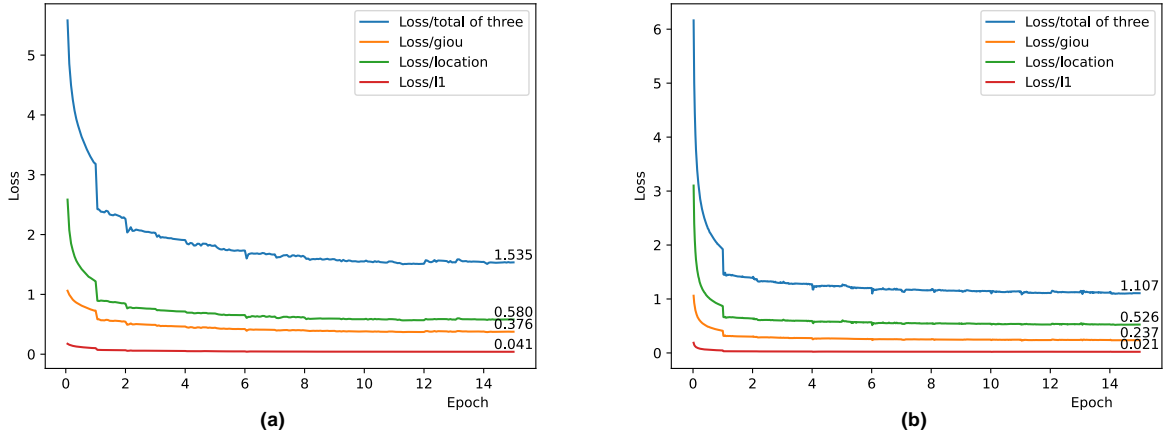

**Figure 1** Comparison of training loss between our proposed student model and FOST. (a) Student model (b) FOST.

### A.2 Why Chose Distillation?

In order to further verify the effectiveness of our proposed distillation strategy, we set up a control experiment, which uses the same model architecture as our proposed student model, but without the hierarchical distillation strategy. Instead, it is directly supervised by the LasHeR dataset. We denote this control model as FOST (Fused One-Stream Track). It is worth mentioning that we used the same optimizer parameters for training FOST as for training the student model. As for the loss function, we removed the distillation-related loss terms $\mathcal{L}_{MF}$ and $\mathcal{L}_{RM}$. Under the same batchsize, we trained the two models for 15 epochs and recorded the change trends of their common loss terms: GIoU Loss, L1 Loss, Location Loss, and the sum of the three. As shown in Figure 1, with the introduction of the hierarchical distillation strategy, the model can mimic the feature distribution of the teacher model and achieve fitting more quickly. (The comparison of PR/SR is already shown in Table 4 of the main text.)

Additionally, we visualized the attention matrices in the backbones of the student model and FOST to demonstrate that the models can effectively locate targets under different dominant advantageous modalities.

For the two different modalities, we calculated the attention values from the search area queries to the template area keys separately, and took the average across all attention heads. As shown in Figure 2, the first four rows represent sequence 1, which involves low-light conditions, favoring the infrared modality; the last four rows represent sequence 2, which involves thermal crossover, favoring the visible modality. By observing the attention maps, we can find that our proposed model can effectively utilize the advantageous modality to achieve target localization and suppress the noise from the disadvantageous modality, regardless of which modality is dominant. Comparing the score maps, we can also observe that our proposed model has more precise localization and smaller noise compared to FOST, further demonstrating that our proposed hierarchical distillation strategy can help the model rapidly learn the complementary characteristics between modalities with relatively small training data, alleviating the training difficulties arising from the increased token length.

## B MORE EVALUATION DATA

This section mainly provides supplementary information regarding the evaluation on public datasets, in order to offer more detailed references.

### B.1 Attribute-Based Evaluation Data

As shown in Tables 3 and 4, we provide detailed data for the attribute-based evaluation on the LasHeR dataset and the RGBT234 dataset. There are 19 challenge attribute sequences in the LasHeR data set, which are: no occlusion (NO), partial occlusion (PO), hyaline occlusion (HO), total occlusion (TO), out-of-view (OV), low illumination (LI). , high illumination (HI), abrupt illumination variation (AIV), low resolution (LR), thermal crossover (TC), deformation (DEF), similar appearance (SA), fast motion (FM), scale variation (SV) ), motion blur (MB), camera shift (CM), aspect ratio change (ARC), frame loss (FL), and background clutter (BC). There are similar challenge attribute annotations in the RGBT234 data set, but there are only 12 types (already included in the above 19 types).

In Table 3, we can observe that our proposed method surpasses most RGB-T trackers in terms of success rate, but lags behind TBSI in precision rate for certain attributes. This indicates that while our model excels in overall tracking capability, there is still room for improvement in tracking precision. On the larger-scale and more challenging LasHeR dataset, as shown in Table 4, our proposed method achieves a more comprehensive lead compared to other RGB-T trackers, further demonstrating its effectiveness.

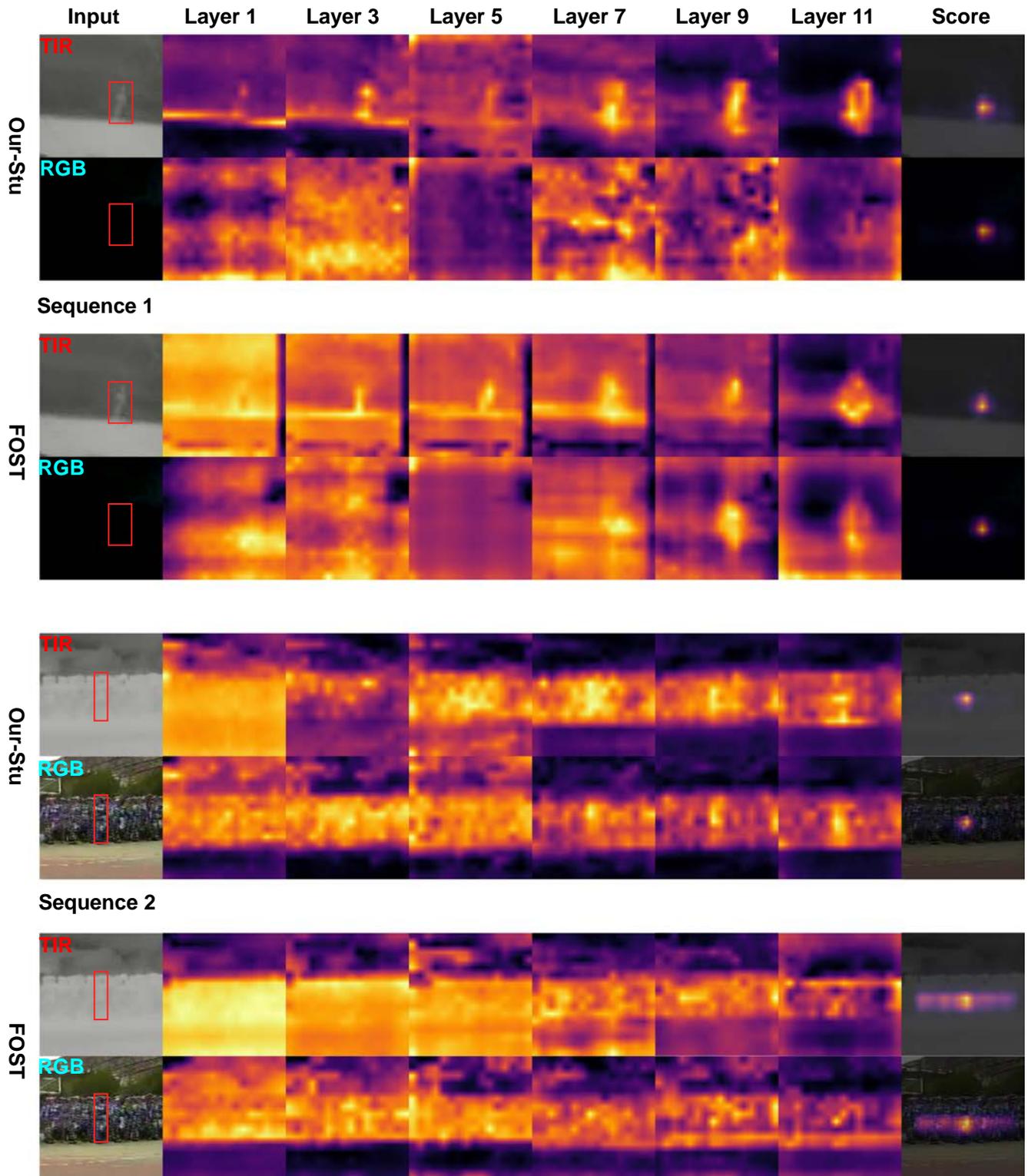

Figure 2 Visualization of the attention maps and response score maps of our proposed student model and FOST.

|    | DMCNet[4] | | ViPT[15] | | APFNet[8] | | MaCNet[12] | | TBSI[1] | | JMMAC[13] | | Ours-Teacher | | Ours-Student | |
|----|----|----|----|----|----|----|----|----|----|----|----|----|----|----|----|----|
|    | PR | SR | PR | SR | PR | SR | PR | SR | PR | SR | PR | SR | PR | SR | PR | SR |
| NO | 93.7/70.6 | | 94.5/74.2 | | 96.0/70.5 | | 94.0/68.4 | | 96.1/72.8 | | 91.6/70.2 | | 98.0/77.5 | | **98.3**/**77.8** | |
| PO | **89.9**/63.0 | | 84.4/63.3 | | 85.8/60.4 | | 81.1/56.5 | | 88.7/64.7 | | 79.9/59.4 | | 88.2/67.1 | | 88.4/**67.3** | |
| HO | 71.5/51.2 | | 77.1/57.2 | | 75.0/51.6 | | 71.9/48.5 | | **81.5**/58.6 | | 67.8/47.6 | | 80.1/**60.3** | | 76.2/57.2 | |
| LI | 85.3/59.2 | | 79.1/57.2 | | 86.8/58.5 | | 81.4/54.0 | | 89.2/63.6 | | 88.4/61.8 | | **89.3**/**66.2** | | 88.2/65.8 | |
| LR | 83.2/58.1 | | 84.6/61.3 | | 83.8/57.3 | | 77.9/53.1 | | **85.1**/60.0 | | 75.3/51.4 | | 83.0/**61.6** | | 82.3/61.2 | |
| TC | 87.0/61.9 | | **88.9**/66.2 | | 84.4/59.1 | | 81.5/59.0 | | 85.8/63.2 | | 72.7/50.4 | | 88.4/**67.3** | | 86.8/63.5 | |
| DEF | 75.4/54.9 | | 81.3/63.4 | | 79.4/56.8 | | 73.5/51.4 | | **84.1**/63.7 | | 68.0/51.2 | | 82.6/**65.2** | | 80.4/64.2 | |
| FM | 81.0/53.6 | | **89.7**/**64.8** | | 86.6/55.6 | | 80.9/49.2 | | 81.4/58.7 | | 68.0/36.6 | | 82.6/62.0 | | 78.6/57.2 | |
| SV | 82.3/59.8 | | 82.6/62.9 | | 83.7/58.5 | | 78.3/55.7 | | **89.9**/66.8 | | 81.0/60.6 | | 88.7/**68.4** | | 85.8/66.0 | |
| MB | 79.4/60.3 | | 86.8/67.7 | | 80.1/60.2 | | 75.8/57.4 | | **88.1**/64.9 | | 74.3/56.7 | | 86.3/**68.4** | | 84.8/67.4 | |
| CM | 81.2/59.4 | | 85.4/65.0 | | 81.0/59.1 | | 75.9/54.2 | | **88.0**/65.0 | | 74.2/55.0 | | 86.7/**66.6** | | 83.9/64.6 | |
| BC | 82.1/54.7 | | 83.7/59.1 | | 81.0/54.1 | | 79.6/49.0 | | 83.4/57.8 | | 63.8/44.3 | | **83.9**/**60.1** | | 79.6/57.8 | |
| ALL | 83.9/59.3 | | 83.5/61.7 | | 82.7/57.9 | | 79.0/55.4 | | 87.1/63.7 | | 79.0/57.3 | | **88.3**/**66.1** | | 87.3/65.1 | |

**Table 3 Detailed attribute-based evaluation data in RGBT234 dataset.**

|    | MANet++[3] | | DMCNet[4] | | ViPT[15] | | MaCNet[12] | | TBSI[1] | | Ours-Teacher | | Ours-Student | |
|----|----|----|----|----|----|----|----|----|----|----|----|----|----|----|
|    | PR | SR | PR | SR | PR | SR | PR | SR | PR | SR | PR | SR | PR | SR |
| NO | 65.3/44.9 | | 68.1/47.5 | | 84.1/68.4 | | 69.2/49.2 | | **91.4**/**74.1** | | 90.2/73.1 | | 89.7/72.7 | |
| PO | 49.8/34.3 | | 52.7/38.0 | | 62.5/50.4 | | 51.7/37.3 | | 67.8/54.0 | | 68.7/54.4 | | **68.8**/**54.5** | |
| TO | 38.2/26.5 | | 42.6/30.9 | | 57.7/46.2 | | 42.8/30.7 | | 64.3/51.0 | | 65.6/51.7 | | **66.3**/**52.0** | |
| HO | 25.2/25.3 | | 29.6/29.8 | | 46.8/43.4 | | 32.1/32.8 | | **60.6**/**53.4** | | 57.4/51.3 | | 60.4/53.0 | |
| OV | 39.2/23.5 | | 46.9/31.7 | | 76.2/65.0 | | 45.3/31.9 | | 64.6/55.9 | | **76.6**/**65.1** | | 72.2/62.0 | |
| LI | 44.1/30.3 | | 48.1/34.8 | | 56.7/41.2 | | 45.2/32.9 | | 61.3/**49.3** | | 60.8/48.5 | | **61.6**/49.1 | |
| HI | 56.8/37.9 | | 61.5/42.8 | | 67.8/54.2 | | 59.1/41.4 | | 73.8/58.2 | | 75.6/59.2 | | **76.4**/**60.2** | |
| AIV | 41.1/29.1 | | 47.6/37.0 | | 36.3/34.2 | | 44.0/32.8 | | **58.2**/**49.8** | | 56.6/49.2 | | 57.4/49.3 | |
| LR | 52.7/31.6 | | 56.0/35.3 | | 56.8/41.8 | | 53.2/33.3 | | 63.9/**47.3** | | 63.9/47.1 | | **64.2**/47.1 | |
| DEF | 43.2/32.8 | | 47.3/37.3 | | 67.6/55.8 | | 44.8/36.2 | | 71.6/58.7 | | 72.5/58.7 | | **75.6**/**61.1** | |
| BC | 45.6/32.2 | | 46.7/34.9 | | 65.0/51.9 | | 46.3/34.4 | | 69.9/**55.7** | | **70.4**/55.4 | | 70.2/55.4 | |
| SA | 48.2/32.8 | | 51.1/36.7 | | 57.4/46.5 | | 49.5/35.6 | | 62.2/50.2 | | 62.0/49.6 | | **63.2**/**50.5** | |
| TC | 45.5/30.9 | | 49.2/34.7 | | 57.4/46.0 | | 47.9/34.0 | | 62.6/50.1 | | 63.5/50.4 | | **63.9**/**50.6** | |
| MB | 48.3/31.5 | | 52.5/36.2 | | 57.6/46.0 | | 52.1/36.2 | | 63.1/49.5 | | 63.5/49.9 | | **65.5**/**51.4** | |
| CM | 49.3/34.1 | | 54.1/39.0 | | 62.0/50.0 | | 52.6/37.9 | | 69.5/55.0 | | **69.9**/**55.4** | | 69.8/55.3 | |
| FL | 34.4/20.2 | | 39.3/28.3 | | 59.7/46.9 | | 33.2/22.4 | | 60.9/47.5 | | **63.2**/49.4 | | **63.2**/**49.5** | |
| FM | 47.1/33.1 | | 51.3/37.6 | | 63.2/51.5 | | 49.8/36.6 | | 69.4/55.7 | | 70.1/56.0 | | **70.3**/**56.2** | |
| SV | 51.6/35.1 | | 54.9/38.9 | | 65.0/52.5 | | 54.1/38.6 | | 70.2/56.2 | | 71.1/56.5 | | **71.2**/**56.6** | |
| ARC | 40.2/28.6 | | 43.2/32.5 | | 59.4/49.5 | | 41.0/31.3 | | 64.3/52.5 | | 63.8/51.7 | | **66.3**/**53.7** | |
| ALL | 52.3/35.5 | | 55.7/39.5 | | 65.1/52.5 | | 54.6/39.2 | | 70.5/56.3 | | 71.2/56.5 | | **71.4**/**56.7** | |

**Table 4 Detailed attribute-based evaluation data in LasHeR dataset.**

## C EXTEND TO OTHER MODALITIES

In this section, we will not limit ourselves to the visible light + infrared modalities, but further evaluate the ability of our proposed tracking method to extend to other modalities. Specifically, we choose the visible light + depth modality and the visible light + event modality for evaluation. For the visible light + depth modality, we use the training set of the DepthTrack[9] dataset for training, and evaluate on its test set. For the visible light + event modality, we use the training set of the VisEvent[7] dataset for training, and evaluate on its test set. For the teacher model, we train for 8 epochs in the RGB-D experiment and 15 epochs in the RGB-E experiment. For the student model, we train for 15 epochs in both experiments.

|         | TSDM[14] | DAL[6] | DDiMP[2] | DeT[9] | OSTrack[11] | SPT[16] | ProTrack[10] | ViPT[15] | Ours-teacher | Ours-student |
|---------|----------|--------|----------|--------|-------------|---------|--------------|----------|--------------|--------------|
| Pr      | 0.393    | 0.512  | 0.503    | 0.560  | 0.536       | 0.527   | 0.583        | 0.592    | **0.619**    | 0.582        |
| Re      | 0.376    | 0.369  | 0.469    | 0.506  | 0.522       | 0.549   | 0.573        | 0.596    | **0.611**    | 0.608        |
| F-score | 0.384    | 0.429  | 0.485    | 0.532  | 0.529       | 0.538   | 0.578        | 0.594    | **0.615**    | 0.595        |

**Table 5 Overall performance on the DepthTrack[9] dataset.**

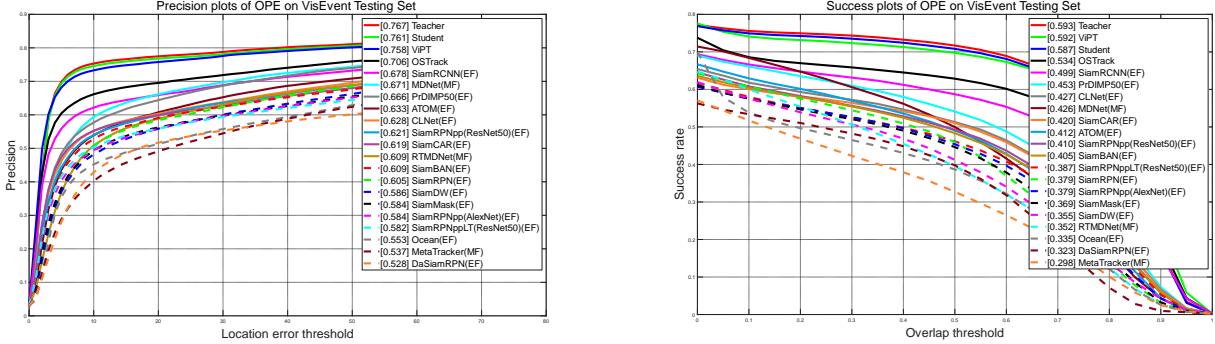

**Figure 3 Overall performance on the VisEvent[7] dataset.**

## C.1 Evaluation On RGB-D Data

The DepthTrack[9] dataset contains 150 sequences for training and 50 sequences for testing. The evaluation metrics include precision (Pr), recall (Re), and F-score. The calculation method of F-score is as follows:

$$F = \frac{2Re \times Pr}{Re + Pr} \quad (3)$$

As shown in Table 5, our proposed teacher model achieved the best performance on both three evaluation metrics. outperforming the baseline[11] by 8.6% in terms of F-score. Meanwhile, our student model also achieved excellent performance on all three evaluation metrics, with an F-score improvement of 6.3% over the baseline[11]. Additionally, in terms of inference speed, the student model achieved an **FPS of 75.8**, while the teacher model had an FPS of 21.5.

## C.2 Evaluation On RGB-E Data

The VisEvent[7] dataset is a very comprehensive RGB-E dataset, with the training set containing 500 sequences and the test set containing 320 sequences.

As shown in Figure 3, our proposed teacher model achieved the best performance on both precision and success metrics, outperforming the baseline[11] by 6.1% in terms of precision. Meanwhile, the performance of the student model is only second to the teacher model, surpassing the baseline[11] by 5.5% in terms of precision. In terms of inference speed, the student model significantly outperforms the teacher model, achieving an **FPS of 90.8**, while the teacher model only achieves 24.2.